\documentclass[runningheads]{llncs}

\usepackage[T1]{fontenc}

\usepackage{graphicx}
\graphicspath{ {./images/} }

\usepackage{hyperref}
\usepackage{color}

\hypersetup{
    colorlinks = true,
    urlcolor = blue,
    linkcolor = blue,
    citecolor = blue
}

\usepackage{multirow}
\usepackage{xcolor}
\usepackage[export]{adjustbox}
\usepackage{colortbl}
\usepackage[caption=false,font=normalsize,labelfont=sf,textfont=sf]{subfig}
\usepackage{amsmath,amssymb,amsfonts}

\begin{document}

\title{Adaptively Augmented Consistency Learning: \\ A Semi-supervised Segmentation Framework \\ for Remote Sensing}


\author{
Hui~Ye\inst{1} 
Haodong~Chen\inst{1}
Xiaoming~Chen\inst{2}
\and
Vera~Chung\inst{1}
}

\authorrunning{H. Ye et al.}

\institute{
University of Sydney, Australia\\ 
\email{huye0731@uni.sydney.edu.au}\\
\email{\{haodong.chen,vera.chung\}@sydney.edu.au} 
\and
Beijing Technology and Business University, China\\
\email{xiaoming.chen@btbu.edu.cn}
}

\maketitle
\begin{abstract}
Remote sensing (RS) involves the acquisition of data about objects or areas from a distance, primarily to monitor environmental changes, manage resources, and support planning and disaster response.
A significant challenge in RS segmentation is the scarcity of high-quality labeled images due to the diversity and complexity of RS image, which makes pixel-level annotation difficult and hinders the development of effective supervised segmentation algorithms.
To solve this problem, we propose Adaptively Augmented Consistency Learning (AACL), a semi-supervised segmentation framework designed to enhances RS segmentation accuracy under condictions of limited labeled data.
AACL extracts additional information embedded in unlabeled images through the use of Uniform Strength Augmentation (USAug) and Adaptive CutMix (AdaCM). 
Evaluations across various RS datasets demonstrate that AACL achieves competitive performance in semi-supervised segmentation, showing up to a 20$\%$ improvement in specific categories and 2$\%$ increase in overall performance compared to state-of-the-art frameworks. 
\end{abstract}

\section{Introduction}
Remote Sensing (RS) technology has revolutionized the observation and analysis of Earth's surface and atmosphere, becoming an indispensable tool with critical applications across various sectors. 
It is widely used in environmental monitoring \cite{Deforestation2019shimabukuro,WaterPollution2020gu}, precise agriculture \cite{PreciseAgriculture2018milioto} and urban planning \cite{UrbanPlaning2017liu}.
In each of these fields, the ability to measure and analyze RS images with high accuracy is essential for effective management and decision-making in real world scenarios. 

Unlike other computer vision tasks, the effectiveness of RS technique is frequently compromised by the scarcity of high-quality labels due to the high resolution and rich information content of RS images. 
Accurate RS techniques heavily depends on the availability of high-quality labels to supervise the model training. 

To overcome the limitations imposed by the scarcity of high-quality labels, we introduce a novel semi-supervised segmentation framework, called Adaptively Augmented Consistency Learning (AACL), to enhance the performance of RS image segmentation.

\begin{figure}[!htp]
    \centering
    \includegraphics[width=0.95\linewidth]{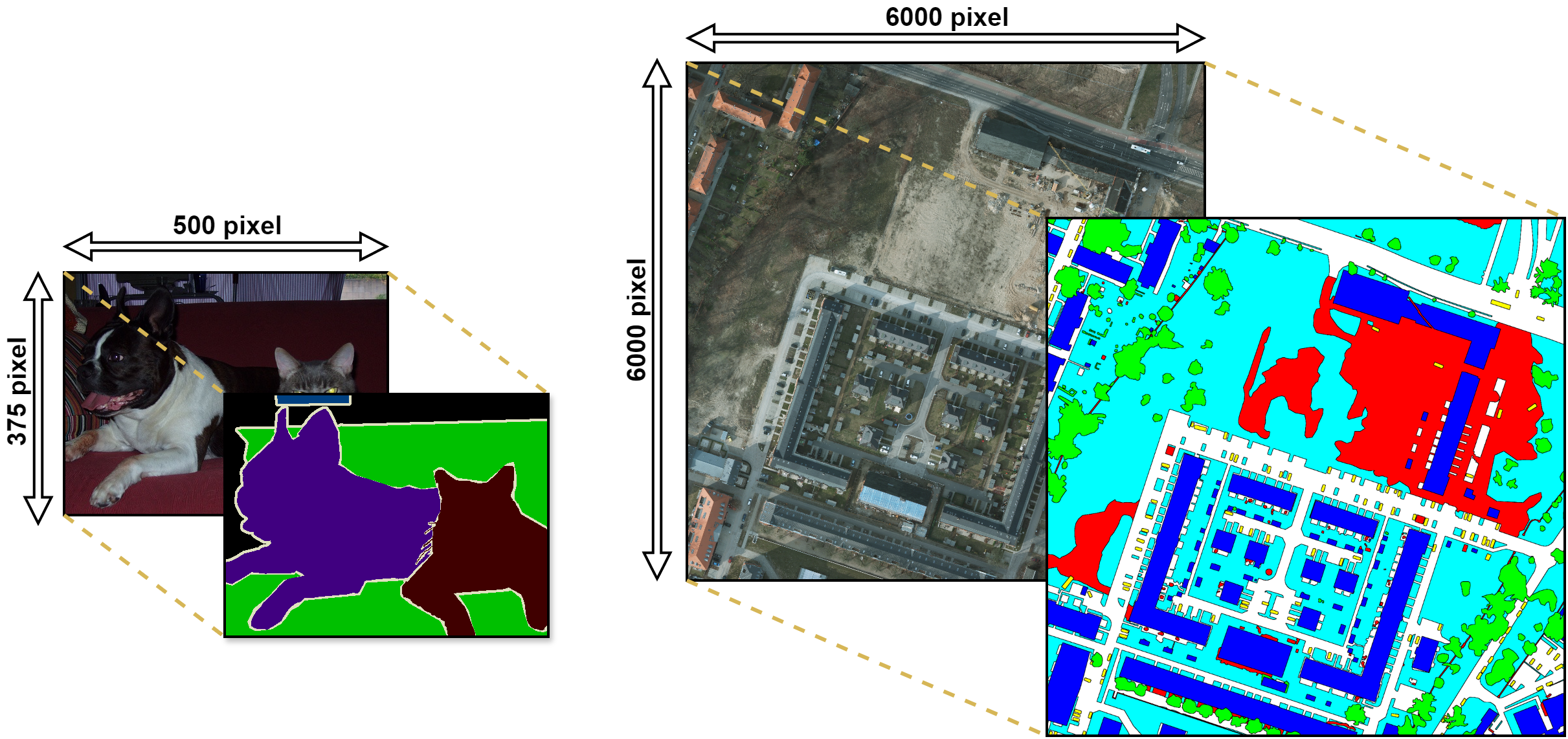}
    \caption{Comparison of natural image and remote sensing image. The image on the left side is from Pascal VOC dataset \cite{Pascal2015everingham}, the image on the right side is from Potsdam dataset.}
\end{figure}

In AACL, two advanced methods, Uniform Strength Augmentation (USAug) and Adaptive CutMix (AdaCM), contribute to the performance improvement.
USAug applies consistent but varied strong augmentation to unlabeled images, introducing discrepancies and enriching the embedded information.
AdaCM tailors the application of the CutMix \cite{CutMix2019yun} based on the model's maturity, dynamically applying CutMix to augment unlabeled images, further introducing discrepancies and mitigating confirmation bias \cite{ConfirmationBias2020arazo}.

The performance of AACL has been validated on multiple mainstream RS datasets, demonstrating notable improvements over existing state-of-the-art (SOTA) semi-supervised segmentation frameworks in RS.

In this paper, we introduce the Adaptively Augmented Consistency Learning (AACL), a novel semi-supervised segmentation framework that significantly advances the efficacy of RS image segmentation. 
This method is specifically designed to address the prevalent issue of scarce high-quality labels, which hampers the accuracy of RS technology. 
The contributions of this work can be summarized as:
\begin{itemize}
    \item we propose Adaptively Augmented Consistency Learning (AACL), a framework designed to enhance RS by effectively leveraging unlabeled images during the training process. This framework addresses the critical challenge of label scarcity in RS, ensuring that the accuracy and reliability of measurements can be maintained with limited labeled data.
    \item We design two novel modules, Uniform Strength Augmentation (USAug) and Adaptive CutMix (AdaCM), which significantly contribute to the performance enhancement of RS image segmentation.
    \item AACL has been rigorously tested and validated on mainstream RS datasets, demonstrating competitive performance and surpassing previous SOTA methods by up to 20$\%$ in specific categories and 2$\%$ in overall performance.
\end{itemize}

\section{Related Work}
\subsection{Semi-supervised Image Segmentation}
\subsubsection{Image Segmentation}
Image segmentation can be viewed as an extension of image classification at a higher dimensional level, where each pixel is classified based on its neighbourhood \cite{Perturbation2020french}. 
This conceptual similarity allows for the adaptation from classification to segmentation. One well-known adaptation is the Fully Convolutional Network (FCN) \cite{FCN2015long}, which replaces the fully connected layers in classification network with convolution layers, enabling pixel-level prediction. Recently, a large segmentation model, called segment anything model (SAM), is proposed  \cite{SAM2023kirillov} to operate across a wide range of datasets and segmentation tasks.

\subsubsection{Semi-supervised Learning}
In semi-supervised learning, the two primary categories are consistency regularization \cite{AugMatter2023zhao,UniMatch2023yang} and self-training \cite{st++2022yang,u2l2022wang}.

In consistency regularization, recent studies emphasize that the strong augmentation technique effective for classification may not always be suitable for segmentation \cite{Perturbation2020french,ClassMix2021olsson}. 
Previous work focused on determining if strong augmentation technique used in classification are effective for segmentation \cite{cutmix2019French} or designing segmentation-specific strong augmentations \cite{ClassMix2021olsson,AugMatter2023zhao}. 
Other approaches introduced multi-level perturbations\cite{StrictMeanTeacher2022liu}, combined consistency regularization with self-training \cite{CPS2021chen}, or design a novel perturbation structures \cite{UniMatch2023yang}.

Self-training methods focused on dynamically adjusting the confidence threshold based on previous training stages \cite{st++2022yang}, extending the refinement stage \cite{GistRist2022teh}, strategically sampling data to address class imbalance issues \cite{RedistributingPL2021he,CSST2021zhu}, and introducing auxiliary tasks \cite{u2l2022wang}.

\subsection{Semi-supervised Remote Sensing Image Segmentation}
Compared with the rapid development in general semi-supervised segmentation, progress in semi-supervised segmentation within the field of RS remains relatively limited. 
However, several notable efforts have been made to enhance the RS application.

Lu et al. \cite{LSST2022lu} introduced a linear-sampling self-training framework that enhances unlabeled data with extra information through strong augmentation and employs class-wise thresholds, optimizing the use of unlabeled data in the training process.
Xin et al. \cite{rsiFeatperturbConstLearn2024xin} developed a segmentation approach that incorporates feature perturbation to expand the perturbation space.
Zhang et al. \cite{JointSTCR2023zhang} introduced an additional branch to FixMatch, enriching the model's ability to learn from varied perspectives and augmentations.

\section{Methodology}
\subsection{Problem Definition}
In the condition of label scarcity, given a small labeled dataset $D_l=\{(x^l_i,y^l_i)\}^{N_l}_{i=1}$ and a larger unlabeled dataset $D_u=\{x^u_i\}^{N_u}_{i=1}$, where $x^l$ and $y^l$ are the labeled image and its corresponding label, respectively, and $x^u$ is the unlabeled image.
Following previous work \cite{wscl2023lu}, $D_l$ will be over-sampled until $N_l \approx N_u$.

The objective function is the combination of supervised loss $\mathcal{L}_s$ and consistency loss $\mathcal{L}_{con}$,
\begin{equation}
\mathcal{L}_{tot} = \mathcal{L}_{s} + \lambda_{con} \mathcal{L}_{con}
\label{eq:ObjFunc}
\end{equation}
where $\lambda_{con}$ is a hyper-parameter that scaling the contribution of $\mathcal{L}_{con}$. Both $\mathcal{L}_s$ and $\mathcal{L}_{con}$ are computed using standard pixel-wise cross-entropy loss $\mathcal{L}_{ce}$,
\begin{equation}
\label{eq:CELoss}
\mathcal{L}_{ce} = - \frac{1}{N} \sum^{N}_{i=1} \sum^C_{c=1}y_{i,c}\log(p_{i,c})
\end{equation}
where $y_{i,c}$ indicates whether class $c$ is the correct classification for pixel $i$, and $p_{i,c}$ is the predicted probability of pixel $i$ being of class $c$.

\begin{figure}[!htp]
    \centering
    \includegraphics[width=0.95\linewidth]{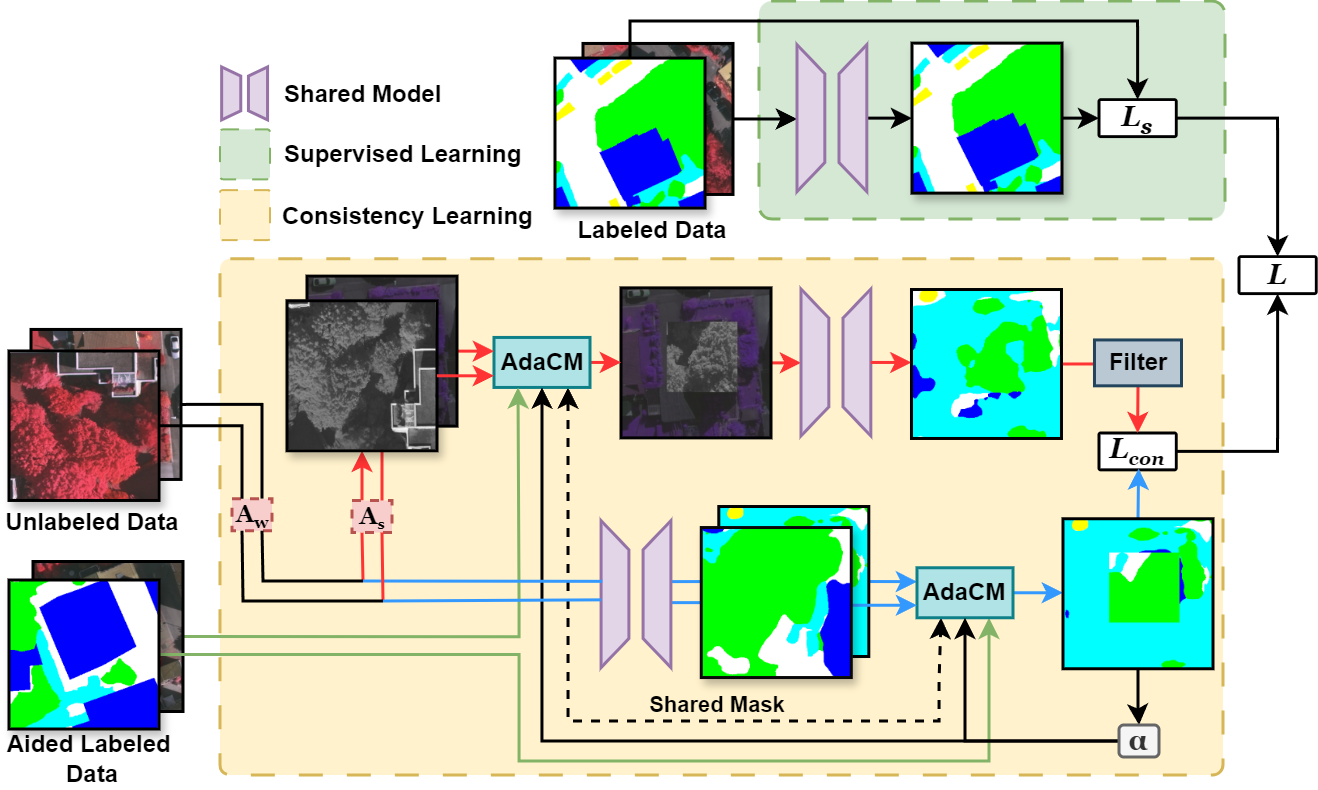}
    \caption{Overview structure of AACL. "$A_w$" and "$A_s$" indicate the weak augmentation and the "USAug" module, respectively.}
    \label{fig:AACL}
\end{figure}

\subsection{Adaptively Augmented Consistency Learning (AACL)}
Fig.~\ref{fig:AACL} illustrates the overall structure of AACL. 
In each iteration, the input consists of two unlabeled images $x^u$ and $x^{u,aux}$, along with two labeled images $x^l$, $x^{l,aid}$ and their corresponding labels $y^l$ and $y^{l,aid}$.

The supervised loss $\mathcal{L}_s$ is simply the cross-entropy loss between the prediction of the labeled image $x^l$ and its label $y^l$,
\begin{equation}
\label{eq:SupLoss}
\mathcal{L}_s = - \frac{1}{N} \sum^{N}_{i=1} \sum^C_{c=1}y^l_{i,c}\log f(x^l_{i,c}; \theta)
\end{equation}
where $f(\cdot; \theta)$ is the shared model with parameter $\theta$.

For consistency loss $\mathcal{L}_{con}$, it is the cross-entropy loss between the prediction of the weakly augmented image $x^w$ and the prediction of strongly augmented image $x^s$,
\begin{equation}
\label{eq:ConLoss}
\mathcal{L}_{con} = - \frac{1}{N} \sum^{N}_{i=1} \sum^C_{c=1}f(x^w_{i,c}; \theta)\log f(x^s_{i,c}; \theta)
\end{equation}

Inspired by previous work \cite{wscl2023lu,u2l2022wang}, to alleviate the adverse impact of unreliable pixels in the strongly augmented prediction $f(x^s; \theta)$, we use entropy, denoted as $H$, as the metrics of the reliability,
\begin{equation}
    \label{eq:entropy}
    H = - \sum^C_{c=1}p_{c}\log p_{c}
\end{equation}
where $p_c$ represents the Softmax probabilities of a pixel in its $c$-th dimension.
Therefore, Eq.(\ref{eq:ConLoss}) can be reformulated as, 
\begin{equation}
    \mathcal{L}_{con} = - \frac{1}{N} \sum^{N}_{i=1} \sum^C_{c=1}f(x^w_{i,c}; \theta)\log f(x^s_{i,c}; \theta) \cdot \mathbb{I} {\{ H_i \leq \tau \}} 
\end{equation}
\begin{equation}
\mathbb{I} {\{ H_i \leq \tau \}}  = 
\begin{cases}
0 & \text{if } H_i \geq \tau,\\
1 & \text{if } H_i < \tau.
\end{cases}
\end{equation}
where $\mathbb{I}{\{ \cdot \}}$ is the indicator function and $\tau$ is the pre-defined threshold. 
This ensures that only data with entropy lower than the threshold $\tau$ contributes to the consistency loss, thereby enhancing the reliability and accuracy of the learning process.

\subsection{Uniform Strength Augmentation}
The advanced performance of the AACL can be attributed to the additional information derived from unlabeled images. 
This perturbation is partly generated by the strong augmentation, which has demonstrated its effectiveness in previous works \cite{LSST2022lu,wscl2023lu,SDAbaseline2021yuan}. 
However, previous works apply strong augmentation either with a fixed order or a fixed type. These approach contradicts the concept of consistency regularization, which emphasizes the variety and multi-dimension of perturbation.

\begin{figure}[!htp]
    \begin{minipage}[c]{.2\textwidth} 
        \centering
        \includegraphics[width=1.1in]{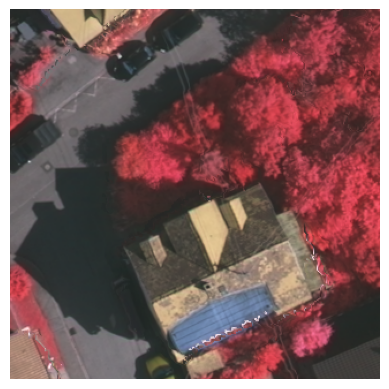}
        \label{fig:main_image}
    \end{minipage}%
    \hfill
    \begin{minipage}[c]{.75\textwidth}
        \centering
        \subfloat[]{\includegraphics[width=0.20\textwidth]{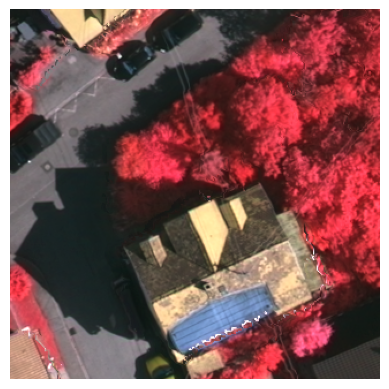}}
        \subfloat[]{\includegraphics[width=0.20\textwidth]{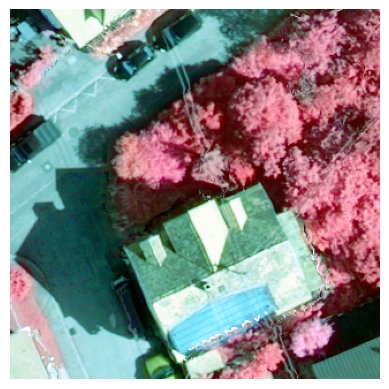}}
        \subfloat[]{\includegraphics[width=0.20\textwidth]{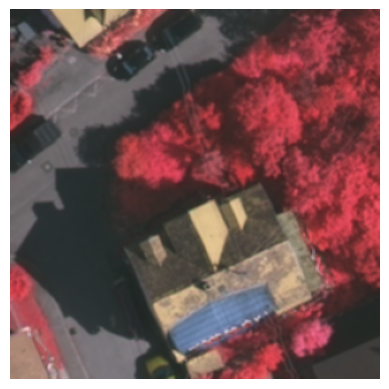}}
        \subfloat[]{\includegraphics[width=0.20\textwidth]{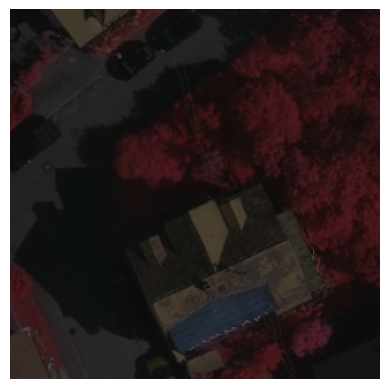}}
        \subfloat[]{\includegraphics[width=0.20\textwidth]{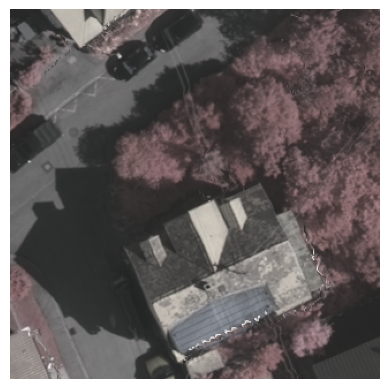}}
        \\
        \subfloat[]{\includegraphics[width=0.20\textwidth]{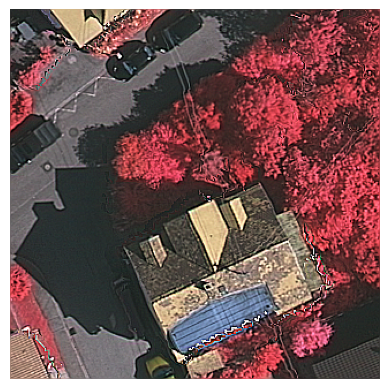}}
        \subfloat[]{\includegraphics[width=0.20\textwidth]{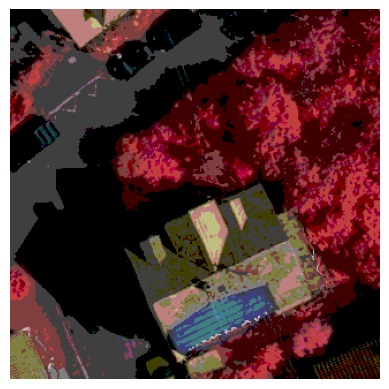}}
        \subfloat[]{\includegraphics[width=0.20\textwidth]{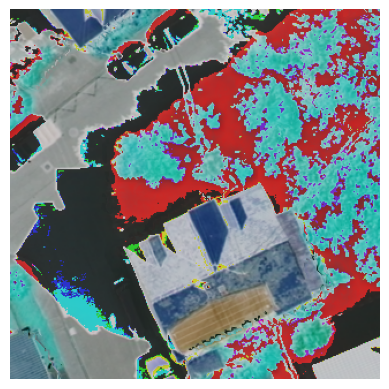}}
        \subfloat[]{\includegraphics[width=0.20\textwidth]{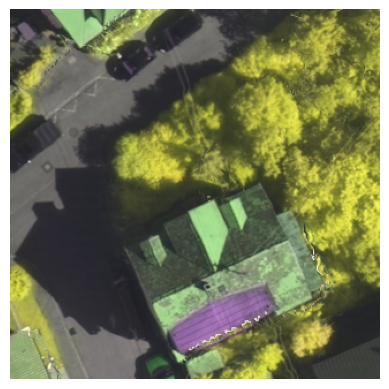}}
        \subfloat[]{\includegraphics[width=0.20\textwidth]{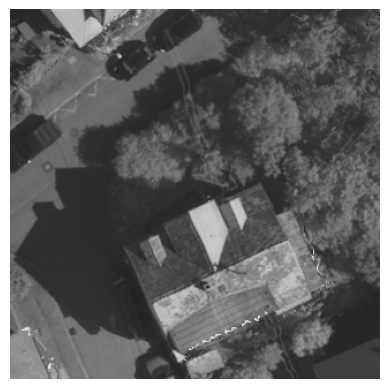}}
    \end{minipage}
    \caption{Visualization of strong augmentation applied in USAug. The image on the left is the original representation, the images on the right are the processed images under different strong augmentation. (a) Contrast. (b) Equalize. (c) Blur. (d) Brightness. (e) Saturation. (f) Sharpness. (g) Posterize. (h) Solarize. (i) Hue. (j) Grayscale.}
    \label{fig:visualization of SDA}
\end{figure}

To address this, we design a Uniform Strength Augmentation (USAug) technique, which applies strong augmentation with varying orders and types but consistent strength. 
USAug selects the commonly used strong augmentations from previous work \cite{wscl2023lu,adaptmatch2023huang,FeatPerturb2024xin} to tailor augmentations for RS images.
The visualization of each strong augmentation is shown in Fig.~\ref{fig:visualization of SDA}. 

To standardize the strength of the strong augmentation, an hyper-parameter $k$ is proposed to represent the number of strong augmentations to be used. 
In each instance of USAug, $k$ augmentations from Fig.~\ref{fig:visualization of SDA} are 
randomly selected and applied in a random order to augment the input image.
This augmentation technique simplifies the hyper-parameter fine-tune process and enhances the model training process. 

\subsection{Adaptive CutMix}
The CutMix technique has demonstrated its effectiveness in many previous semi-supervised learning frameworks \cite{UniMatch2023yang,wscl2023lu}. 
However, previous works only apply CutMix in unlabeled images, which may introduce confirmation bias \cite{ConfirmationBias2020arazo}, especially when training on RS images that are difficult to handle.
Therefore, we propose a novel CutMix technique called Adaptive CutMix (AdaCM). It applies CutMix either between two unlabeled images or between one labeled image and one unlabeled image, depending on the model's confidence. 

\begin{figure}[!htp]
    \centering
    \includegraphics[width=0.8\linewidth]{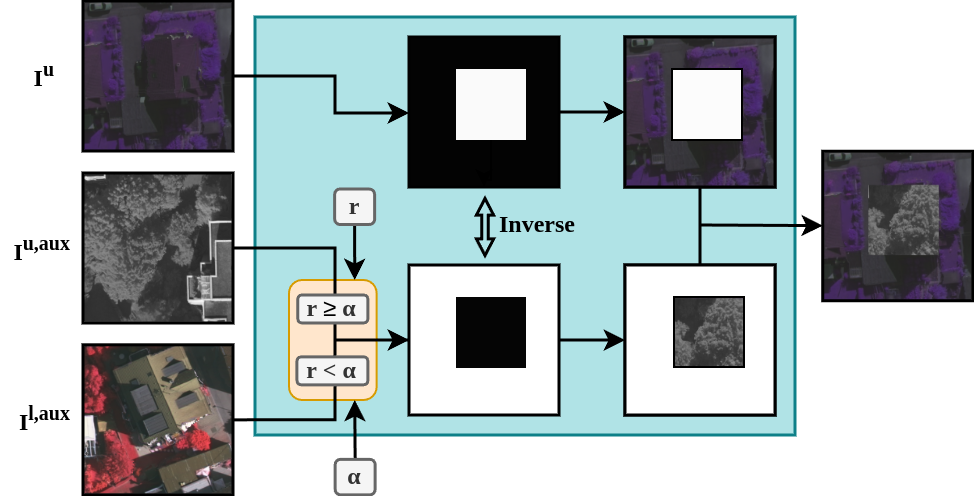}
    \caption{The structure of AdaCM module. $I^u$, $I^{u,aux}$ and $I^{l,aux}$ are unlabeled image, auxiliary unlabeled image and auxiliary labeled image, respectively. "$r$" is the random threshold in AdaCM, "$\alpha$" is the trigger probability from the weakly augmented prediction.}
    \label{fig:AdaCM}
\end{figure}

The logic of AdaCM is as follows: 
At the early stages of training, the model is unreliable and prone to making inaccurate predictions. 
During this phase, CutMix is applied between an unlabeled image and a labeled image to foster more stable learning. 
As the model becomes well-trained and generates more reliable predictions, the application of CutMix shifts towards combining two unlabeled images, introducing further perturbations and enhancing the model's generalization capability. 

The implementation of AdaCM is shown in Fig.~\ref{fig:AdaCM}, and can be formulated as,
\begin{equation}
I = 
\begin{cases}
I^u \odot M_n + I^{u,aux} \odot (1 - M_n) & \text{if } r \geq \alpha,\\
I^u \odot M_n + I^{l,aux} \odot (1 - M_n) & \text{if } r < \alpha.
\end{cases}
\label{eq:AdaCM}
\end{equation}
where $r$ is the random threshold, $M_n$ represents the region CutMix mask, and $\alpha$ is the trigger probability computed by,
\begin{equation}
\label{eq:confidence threshold}
    \alpha = \frac{1}{N}\sum^{N}_{i=1}max(p_{i,c})(1-\frac{-\sum p_{i,c}\log p_{i,c}}{C})
\end{equation}
where $p_{i,c}$ represents the Softmax probabilities of the $i$-th pixel for  class $c$.

It is worth noting that there are two instance of AdaCM in AACL: one for the strongly augmented image and another for the prediction of the weakly augmented image. 
They share the same $r$, $\alpha$ and $M_n$, ensuring the geometric consistency between weakly and strong augmented views.
This dynamic application of CutMix ensures the model progressively refines its augmentation strategy as its prediction confidence improves.

\section{Experiment}
\subsection{Experimental Setup}
\subsubsection{Datasets}
To validate the advancement of AACL, the DFC22 \cite{DFC22_2022hansch}, iSAID \cite{iSAID2019waqas} and Vaihingen datasets are selected for this work. Only single data partition is applied for fair comparison with previous SOTA framework \cite{wscl2023lu}.

The DFC22 \cite{DFC22_2022hansch} consists of 12 distinct classes and comprises 766 high-resolution labeled images, each with a resolution of approximately are around $2,000 \times 2,000$ pixels. 
In our experiment, a random selection of $80\%$ of the images is allocated to the training set, with the remainder used for testing.

The iSAID dataset \cite{iSAID2019waqas} comprises $1,411$ high-resolution images, featuring a wide range of resolutions from $12,029 \times 5,014$ pixels to $455 \times 387$ pixels, and includes 15 distinct categories. 
In our experiment, $1,200$ images are designated as the training set, with the remaining images serving as the test set.

The Vaihingen dataset varies in resolution, ranging from $1,996 \times 1,995$ pixels to $3,816 \times 2,550$. The dataset includes 5 land cover classes: buildings, roads, trees, grass, and cars. 
In the following experiments, 16 patches are selected as the training set, and the remaining 16 patches serve as the testing set.

\subsubsection{Implementation Setting}
For the segmentation model, we employ Deeplabv3+ \cite{deeplabv3p2018chen} with a ResNet-101 backbone \cite{ResNet2016he}. 
All the experiments are conducted on a server equipped with an 16GB NVIDIA 4080 GPU.

For training across all datasets, the batch size is maintained at 16, comprising an equal split of 8 labeled images and 8 unlabeled images. 
The optimizer employed is Stochastic Gradient Descent (SGD) with a base learning rate of $1e^{-3}$, a weight decay of $1e^{-4}$, and momentum of $0.9$. 
The number of training epochs is $100$ for all datasets, with both training and testing images sized at $321 \times 321$ pixels. 

The scaling factor of the consistency loss, $\lambda_{con}$, is set to 1 because $\lambda_{con}$ is empirically proven \cite{wscl2023lu,AugMatter2023zhao} to have minimal impact on overall performance.
The threshold parameter $\tau$ is set to $20$ for both DFC22 and iSAID datasets and adjusted to $80$ for the Vaihingen dataset, following previous work \cite{wscl2023lu}.

\subsection{Experiment Result and Analysis}
\subsubsection{Results on DFC22 Dataset.}
Table~\ref{tab:DFC22 result} presents a comparative analysis of our method against the baseline and other leading techniques on the DFC22 dataset.

Compared to a supervised-only baseline, our proposed semi-supervised framework significantly enhances mean Intersection over Union (mIoU), demonstrating performance boosts of 11.69$\%$ and 6.6$\%$ for the $1/8$ and $1/4$ data partition schemes, respectively. 

\begin{table}[!ht]
\centering
\caption{Per-class IoU and mean IoU on DFC22 Dataset. The best results are in bold while the second best results are underlined (the same for the rest tables).}
\label{tab:DFC22 result}
\resizebox{\textwidth}{!}{%
\begin{tabular}{ccccccccccccccc}
\hline
Labeled & Method & Urban & Industrial & Mine & Artificial & Arable & Permanent & Pasture & Forest & Herbaceous & Open & Wetland & Water & \textbf{mIoU} \\ \hline
\multirow{9}{*}{1/8} & Baseline & 54.10 & 29.54 & 0.81 & 7.27 & 10.03 & 43.37 & 46.84 & 61.19 & 32.88 & 22.07 & 8.20 & 7.35 & 26.97 \\
 & AdvSS & 58.63 & 29.55 & 0.08 & 7.21 & 7.44 & 49.33 & 52.42 & 59.68 & 36.35 & 24.48 & 5.27 & 4.87 & 27.94 \\
 & s4GAN & 58.67 & 28.12 & 0.09 & 10.17 & \textbf{12.56} & 55.37 & 44.85 & 64.76 & 37.01 & 18.42 & 0.09 & 16.96 & 28.92 \\
 & ClassMix & 56.77 & 29.68 & 3.32 & 1.06 & 2.69 & 36.92 & 53.91 & 65.35 & 36.17 & 0.38 & 3.16 & 4.59 & 24.50 \\
 & SS-Cont & 58.12 & 29.86 & NAN & 0.38 & 9.67 & NAN & 49.62 & 55.14 & 11.00 & 0.01 & 0.46 & NAN & 17.86 \\
 & ST++ & 60.18 & 33.58 & NAN & 15.39 & \underline{11.99} & \textbf{59.95} & \underline{54.25} & 63.72 & 35.58 & 1.86 & 0.83 & 15.62 & 29.41 \\
 & LSST & 60.54 & 38.44 & \underline{5.11} & \textbf{17.15} & 9.76 & 52.29 & 52.81 & 63.51 & 36.11 & 5.49 & 6.55 & 23.53 & 30.94 \\
 & WSCL & \underline{61.19} & \textbf{40.66} & \textbf{5.50} & \underline{16.16} & 10.35 & 58.13 & 53.14 & \textbf{70.32} & \textbf{40.44} & \underline{38.26} & \underline{31.58} & \textbf{30.34} & \underline{38.00} \\ 
 & \cellcolor[HTML]{D3D3D3}{AACL (Ours)} & \cellcolor[HTML]{D3D3D3}{\textbf{63.27}} & \cellcolor[HTML]{D3D3D3}{\underline{40.39}} & \cellcolor[HTML]{D3D3D3}{1.11} & \cellcolor[HTML]{D3D3D3}{12.47} & \cellcolor[HTML]{D3D3D3}{10.86} & \cellcolor[HTML]{D3D3D3}{\underline{58.37}} & \cellcolor[HTML]{D3D3D3}{\textbf{55.25}} & \cellcolor[HTML]{D3D3D3}{\underline{66.61}} & \cellcolor[HTML]{D3D3D3}{\underline{37.70}} & \cellcolor[HTML]{D3D3D3}{\textbf{40.28}} & \cellcolor[HTML]{D3D3D3}{\textbf{51.75}} & \cellcolor[HTML]{D3D3D3}{\underline{25.83}} & \cellcolor[HTML]{D3D3D3}{\textbf{38.66}} \\ \hline
\multirow{9}{*}{1/4} & Baseline & 55.09 & 31.12 & \textbf{1.34} & 5.58 & 7.26 & 51.21 & 51.22 & 65.80 & 37.65 & 24.58 & 51.36 & 9.88 & 32.67 \\
 & AdvSS & 61.30 & 35.57 & 0.11 & 4.58 & 3.76 & 51.92 & \underline{57.48} & 65.90 & 38.44 & 14.57 & 51.42 & 14.44 & 33.29 \\
 & s4GAN & 59.85 & 32.82 & NAN & 2.75 & 4.13 & 50.08 & 54.23 & 65.34 & 36.35 & 10.23 & 54.09 & 15.08 & 32.08 \\
 & ClassMix & 55.97 & 32.19 & \underline{0.15} & 3.05 & 5.19 & 23.64 & 50.75 & 66.05 & 17.08 & 13.97 & 40.50 & 11.30 & 26.65 \\
 & SS-Cont & \underline{63.82} & 33.99 & NAN & 4.05 & 3.12 & NAN & 55.08 & 63.75 & 0.09 & NAN & 15.08 & NAN & 9.92 \\
 & ST++ & 59.27 & 38.55 & NAN & 1.65 & 6.42 & 51.17 & 57.32 & \underline{69.90} & 40.01 & 12.53 & 54.18 & 18.53 & 34.12 \\
 & LSST & 61.86 & 36.14 & NAN & 7.48 & \underline{8.26} & 53.28 & 55.65 & 67.97 & 40.72 & 19.37 & \underline{61.24} & \underline{24.80} & 36.92 \\
 & WSCL & 62.47 & \textbf{41.29} & 0.02 & \underline{13.84} & \textbf{12.62} & \underline{54.94} & 50.81 & \textbf{71.04} & \textbf{44.85} & \underline{39.56} & 50.40 & \textbf{25.25} & \underline{38.92} \\ 
 & \cellcolor[HTML]{D3D3D3}{AACL (Ours)} & \cellcolor[HTML]{D3D3D3}{\textbf{64.29}} & \cellcolor[HTML]{D3D3D3}{\underline{41.21}} & \cellcolor[HTML]{D3D3D3}{0.00} & \cellcolor[HTML]{D3D3D3}{\textbf{14.08}} & \cellcolor[HTML]{D3D3D3}{3.96} & \cellcolor[HTML]{D3D3D3}{\textbf{57.82}} & \cellcolor[HTML]{D3D3D3}{\textbf{58.82}} & \cellcolor[HTML]{D3D3D3}{69.18} & \cellcolor[HTML]{D3D3D3}{\underline{41.55}} & \cellcolor[HTML]{D3D3D3}{\textbf{48.97}} & \cellcolor[HTML]{D3D3D3}{\textbf{61.65}} & \cellcolor[HTML]{D3D3D3}{9.76} & \cellcolor[HTML]{D3D3D3}{\textbf{39.27}} \\ \hline
\end{tabular}%
}
\end{table}

Furthermore, when compared to the previous SOTA, our framework achieves marginal improvements in mIoU, with increases of 0.66$\%$ and 0.35$\%$ under $1/8$ and $1/4$ partition protocol, respectively. 
Remarkably, our approach achieves significant increases in 'Wetland' category, with improvements of 20.17$\%$ and 11.25$\%$ under $1/8$ and $1/4$ partition protocol, respectively, demostrating the superior capability of our framework in recognizing 'Wetland'.
Additionally, our framework also achieves increases of 2.11$\%$ and 8.01$\%$ in Intersection over Union (IoU) for the 'Pasture' category under the different partition protocols.
Moreover, our framework also surpass previous SOTA in 'Open' category, with increments of 2.02$\%$ and 9.41$\%$ in IoU under $1/8$ and $1/4$ partition protocol, respectively.

In general, compared with previous SOTA, our framework shows better improvement under $1/4$ partition protocol in the DFC22 dataset. 

\subsubsection{Results on iSAID Dataset}
Table~\ref{tab:iSAID result} displays a comparative analysis of our framework, the baseline, and another advanced framework on the iSAID dataset.

Compared to the baseline, our framework exhibits a substantial enhancement in mIoU, with increases of 20.51$\%$ and 11.95$\%$ under $1/8$ and $1/4$ partition protocols, respectively. Our method achieves similar performance to previous SOTA in 'HC' category, which is hardly recognized under a supervised baseline and even for some of the advanced frameworks.

\begin{table*}[!ht]
\centering
\caption{Per-class IoU and mean IoU on iSAID Dataset. (ST: Storage Tank, BD: Baseball Diamond, TC: Tennis Court, BC: Basketball Court, GTF: Ground Track Field, LV: Large Vehicle, SV: Small Vehicle, HC: HeliCopter, SP: Swimming Pool, RA: roundabout, SBF: Soccer Ball Field)}
\label{tab:iSAID result}
\resizebox{0.95\textwidth}{!}{%
\begin{tabular}{cccccccccccccccccc}
\hline
Labeled & Method & Ship & ST & BD & TC & BC & GTF & Bridge & LV & SV & HC & SP & RA & SBF & Plane & Harbor & \textbf{mIoU} \\ \hline
\multirow{8}{*}{100} & Baseline & 37.28 & 4.70 & 62.30 & 80.00 & 4.95 & 38.85 & 1.16 & 66.52 & 46.76 & NAN & 63.24 & \textbf{65.60} & 48.66 & 38.27 & 40.31 & 39.91 \\
 & AdvSS & 38.46 & 3.95 & 60.24 & 73.92 & 6.37 & 34.55 & 0.32 & 66.32 & 43.56 & NAN & 46.36 & 5.52 & 43.16 & 39.01 & 46.87 & 33.91 \\
 & s4GAN & 40.52 & 0.79 & 48.09 & 75.52 & 9.54 & 34.61 & 0.38 & 72.13 & 44.29 & NAN & 47.47 & 48.71 & 48.30 & 28.08 & 45.74 & 36.28 \\
 & ClassMix & 28.59 & 1.59 & 48.32 & 72.45 & 7.76 & 44.00 & \underline{7.97} & 68.62 & 38.07 & NAN & 60.13 & 52.74 & 43.37 & 28.77 & 45.05 & 36.49 \\
 & ST++ & 45.85 & NAN & 52.44 & 83.38 & 10.79 & 22.60 & NAN & 73.52 & 48.45 & NAN & 49.05 & 37.88 & 59.90 & 46.28 & 48.75 & 38.59 \\
 & LSST & \underline{46.83} & 1.85 & \underline{73.46} & 84.69 & 15.01 & 42.44 & 2.09 & \underline{77.25} & 54.01 & 10.60 & 70.00 & \underline{51.37} & 60.42 & 62.82 & 51.30 & 46.94 \\
 & WSCL & 39.17 & \textbf{87.31} & 59.59 & \textbf{90.93} & \textbf{75.27} & \underline{62.59} & \textbf{14.81} & 76.36 & \underline{56.85} & \textbf{22.89} & \textbf{74.07} & 35.24 & \underline{62.13} & \textbf{80.52} & \underline{56.26} & \underline{59.60} \\
 & \cellcolor[HTML]{D3D3D3}{AACL (Ours)} & \cellcolor[HTML]{D3D3D3}{\textbf{52.45}} & \cellcolor[HTML]{D3D3D3}{\underline{79.94}} & \cellcolor[HTML]{D3D3D3}{\textbf{79.95}} & \cellcolor[HTML]{D3D3D3}{\underline{90.84}} & \cellcolor[HTML]{D3D3D3}{\underline{59.88}} & \cellcolor[HTML]{D3D3D3}{\textbf{63.93}} & \cellcolor[HTML]{D3D3D3}{2.10} & \cellcolor[HTML]{D3D3D3}{\textbf{78.15}} & \cellcolor[HTML]{D3D3D3}{\textbf{57.39}} & \cellcolor[HTML]{D3D3D3}{\underline{21.49}} & \cellcolor[HTML]{D3D3D3}{\underline{72.57}} & \cellcolor[HTML]{D3D3D3}{36.63} & \cellcolor[HTML]{D3D3D3}{\textbf{72.71}} & \cellcolor[HTML]{D3D3D3}{\underline{74.92}} & \cellcolor[HTML]{D3D3D3}{\textbf{63.34}} & \cellcolor[HTML]{D3D3D3}{\textbf{60.42}} \\ \hline
\multirow{8}{*}{300} & Baseline & 45.03 & 83.79 & 65.59 & 93.11 & 66.75 & 61.90 & 25.70 & 75.42 & 55.63 & 13.42 & 71.78 & \underline{58.82} & 66.12 & 72.58 & 52.39 & 60.47 \\
 & AdvSS & 45.04 & 42.00 & 76.10 & 92.90 & 73.20 & 53.01 & 1.88 & 72.42 & 51.39 & NAN & 56.41 & 13.88 & 45.17 & 49.02 & 51.17 & 48.24 \\
 & s4GAN & 45.41 & 77.95 & 65.86 & 91.77 & 59.24 & 53.31 & 15.82 & 73.00 & 55.70 & NAN & 59.42 & 37.55 & 65.92 & 65.35 & 52.48 & 54.59 \\
 & ClassMix & 49.02 & 73.18 & 73.13 & 81.64 & 47.12 & 59.84 & 10.64 & 73.47 & 41.08 & NAN & 66.25 & 52.95 & 36.90 & 56.68 & 46.29 & 51.21 \\
 & ST++ & 37.92 & 86.18 & 68.46 & 95.24 & 63.64 & 52.21 & \underline{27.35} & 77.05 & 55.51 & NAN & 63.49 & 58.38 & 67.55 & 72.19 & 51.57 & 58.45 \\
 & LSST & \underline{52.93} & \textbf{88.11} & 79.63 & 96.95 & 82.52 & 53.28 & 26.41 & 76.59 & 62.54 & 1.01 & 71.65 & 53.62 & 74.63 & \textbf{77.64} & 53.55 & 63.40 \\
 & WSCL & 50.26 & \underline{86.56} & \textbf{91.99} & \underline{97.19} & \textbf{91.26} & \textbf{68.26} & \textbf{28.11} & \underline{81.26} & \textbf{68.71} & \textbf{73.89} & \textbf{82.05} & 54.22 & \textbf{80.17} & \underline{76.23} & \underline{54.76} & \underline{72.33} \\
 & \cellcolor[HTML]{D3D3D3}{AACL (Ours)} & \cellcolor[HTML]{D3D3D3}{\textbf{54.75}} & \cellcolor[HTML]{D3D3D3}{75.54} & \cellcolor[HTML]{D3D3D3}{\underline{86.39}} & \cellcolor[HTML]{D3D3D3}{\textbf{97.70}} & \cellcolor[HTML]{D3D3D3}{\underline{90.71}} & \cellcolor[HTML]{D3D3D3}{\underline{67.68}} & \cellcolor[HTML]{D3D3D3}{23.94} & \cellcolor[HTML]{D3D3D3}{\textbf{81.72}} & \cellcolor[HTML]{D3D3D3}{\underline{68.03}} & \cellcolor[HTML]{D3D3D3}{\underline{72.53}} & \cellcolor[HTML]{D3D3D3}{\underline{78.50}} & \cellcolor[HTML]{D3D3D3}{\textbf{71.37}} & \cellcolor[HTML]{D3D3D3}{\underline{78.72}} & \cellcolor[HTML]{D3D3D3}{74.70} & \cellcolor[HTML]{D3D3D3}{\textbf{63.99}} & \cellcolor[HTML]{D3D3D3}{\textbf{72.42}} \\ \hline
\end{tabular}%
}
\end{table*}

Compared to the previous SOTA \cite{wscl2023lu}, our proposed framework shows incremental improvements in mIoU by 0.82$\%$ and 0.09$\%$ under $1/8$ and $1/4$ partition protocol, respectively. 
Specifically, our framework achieves increments of 13.28$\%$ and 4.49$\%$ increments in the 'Ship' category under the $100$ and $300$ partition protocols. 
Additionally, there are increments of 1.39$\%$ and 17.15$\%$ in 'Round About (RA)' under the $100$ and the $300$ partition protocols. 
Furthermore, increments of 7.08$\%$ and 9.23$\%$ are observed in the 'Harbor' category under the $100$ and $300$ partition protocols.

\subsubsection{Results on Vaihingen Dataset}
Table~\ref{tab:Vaihingen result} provides a comparative analysis of our proposed framework, the baseline, and other leading frameworks within the Vaihingen dataset.

Compared to the supervised-only baseline, AACL demonstrates substantial enhancements in mIoU, with increases of 15.7$\%$ and 11.69$\%$ observed under $1/8$ and $1/4$ partition protocols, respectively. 
Furthermore, AACL surpasses the performance of the previous SOTA framework by 2.35$\%$ and 2.18$\%$, under the $1/8$ and $1/4$ partition protocols, respectively. 
A particular highlight of AACL's performance is observed in the 'Car' category, where it achieves a remarkable improvement in mIoU of 11.92$\%$ and 8.04$\%$ for the $1/8$ and $1/4$ partition protocols, respectively.

\begin{table}[!ht]
\centering
\caption{Per-class IoU and mean IoU on Vaihingen Dataset. (IS: Impervious Surface LV: Low Vegetation)}
\label{tab:Vaihingen result}
\resizebox{0.7\linewidth}{!}{%
\begin{tabular}{cccccccc}
\hline
Labeled & Method & IS & Building & LV & Tree & Car & \textbf{mIoU} \\ \hline
\multirow{9}{*}{1/8} & Baseline & 64.90 & 72.13 & 44.52 & 63.56 & 21.39 & 53.30 \\
 & AdvSS & 67.83 & 74.78 & 50.18 & 67.31 & 25.62 & 57.14 \\
 & s4GAN & 70.77 & 77.78 & 56.28 & 69.03 & 15.37 & 57.85 \\
 & ClassMix & 67.86 & 76.22 & 48.55 & 67.48 & 19.91 & 56.00 \\
 & SS-Cont & 60.99 & 67.98 & 38.84 & 63.55 & 9.87 & 48.25 \\
 & ST++ & 72.71 & 79.16 & 58.97 & 69.22 & 19.52 & 59.92 \\
 & LSST & 73.73 & 79.79 & \underline{59.94} & 70.65 & 36.32 & 64.09 \\
 & WSCL & \textbf{75.69} & \textbf{82.39} & \textbf{61.05} & \underline{70.97} & \underline{43.18} & \underline{66.65} \\
 & \cellcolor[HTML]{D3D3D3}{AACL (Ours)} & \cellcolor[HTML]{D3D3D3}{\underline{75.52}} & \cellcolor[HTML]{D3D3D3}{\underline{81.42}} & \cellcolor[HTML]{D3D3D3}{59.33} & \cellcolor[HTML]{D3D3D3}{\textbf{73.64}} & \cellcolor[HTML]{D3D3D3}{\textbf{55.10}} & \cellcolor[HTML]{D3D3D3}{\textbf{69.00}} \\ \hline
\multirow{9}{*}{1/4} & Baseline & 69.68 & 77.98 & 50.92 & 65.46 & 32.47 & 59.30 \\
 & AdvSS & 73.07 & 81.02 & 54.22 & 68.88 & 40.25 & 63.49 \\
 & s4GAN & 74.28 & 81.19 & 58.97 & 69.87 & 22.85 & 61.43 \\
 & ClassMix & 71.66 & 79.15 & 54.79 & 68.55 & 30.75 & 60.98 \\
 & SS-Cont & 66.99 & 71.75 & 42.85 & 67.05 & 17.69 & 53.22 \\
 & ST++ & 73.56 & 80.71 & 59.46 & 68.71 & 26.18 & 61.72 \\
 & LSST & 75.65 & 82.36 & 61.29 & 69.96 & 37.43 & 65.34 \\
 & WSCL & \underline{77.38} & \textbf{84.61} & \textbf{63.49} & \underline{71.33} & \underline{47.24} & \underline{68.81} \\
 & \cellcolor[HTML]{D3D3D3}{AACL (Ours)} & \cellcolor[HTML]{D3D3D3}{\textbf{77.76}} & \cellcolor[HTML]{D3D3D3}{\underline{83.36}} & \cellcolor[HTML]{D3D3D3}{\underline{62.69}} & \cellcolor[HTML]{D3D3D3}{\textbf{75.35}} & \cellcolor[HTML]{D3D3D3}{\textbf{55.28}} &\cellcolor[HTML]{D3D3D3}{\textbf{70.99}} \\ \hline
\end{tabular}%
}
\end{table}

AACL achieve the most significant improvement in Vaihingen dataset, probably because the effectiveness of AACL relies on the perturbations introduced. 
When the model is fully trained with labeled data, AACL can further enhance performance by leveraging the embedded information in unlabeled data. 
However, when the model has not fully learned the features from labeled data, the benefit of integrating additional information from unlabeled data is relatively limited.

\subsection{Ablation Studies}
\subsubsection{Effectiveness of individual components}
In AACL, the effectiveness of each component is detailed in Table~\ref{tab:ablation component}.
The experiments are conducted under $1/4$ partition protocol on the Vaihingen dataset.
In this ablation study, when "USAug" is not employed, a fixed strong augmentation from prior work \cite{wscl2023lu} is applied.
Similarly, when "AdaCM" is not used, traditional CutMix between two unlabeled images is implemented.

\begin{table}[!h]
\centering
\caption{Effectiveness of each component.}
\label{tab:ablation component}
\resizebox{0.25\linewidth}{!}{%
\begin{tabular}{ccc}
\hline
USAug & AdaCM & mIoU \\ \hline
 & & 69.64 \\
\checkmark & & 70.50 \\
 & \checkmark & 70.46 \\
\checkmark & \checkmark & \textbf{70.99} \\ \hline
\end{tabular}
}
\end{table}

The experiment results reveal that, compared to a baseline model utilizing only fix augmentation and traditional CutMix, the individual contributions of our proposed USAug and AdaCM module to performance improvements are 0.86$\%$ and 0.82$\%$, respectively. 
Furthermore, when both modules are applied together, the performance enhancement reaches 1.45$\%$, indicating a synergistic effect that significantly surpasses the performance improvement offered by each module independently.

\subsubsection{Impact of Augmentation Strength}
we also conduct experiments to assess the effectiveness of the number of augmentations, and the results are presented in Table~\ref{tab:ablation USAug}. 
For this ablation study, a $1/4$ partition protocol is adopted for both DFC22 and Vaihingen datasets, whereas the iSAID dataset adopt $300$ partition protocol. 
To maintain consistency in overall perturbation, traditional CutMix was applied across all datasets. 

\begin{table}[!h]
\centering
\caption{Effectiveness of USAug with Different Augmentation Number $k$}
\label{tab:ablation USAug}
\resizebox{0.35\linewidth}{!}{%
\begin{tabular}{c|ccc}
\hline
k & DFC22 & iSAID & Vaihingen \\ \hline
1 & 31.84 & 61.34 & 68.52 \\
2 & 29.85 & 64.51 & 68.49 \\
3 & \textbf{33.63} & 61.48 & 69.70 \\
4 & 32.96 & 60.71 & 69.67 \\
5 & 32.23 & 62.09 & 69.68 \\
6 & 32.07 & 59.64 & 70.25 \\
7 & 32.16 & 59.10 & 70.30 \\
8 & 31.39 & \textbf{64.70} & \textbf{70.99} \\
9 & 33.02 & 60.30 & 70.07 \\
10 & 33.12 & 60.87 & 69.81 \\ \hline
\end{tabular}
}
\end{table}

The results indicate that the model achieved its peak performance when $k=3$ for the DFC22 dataset. 
iSAID and Vaihingen datasets demonstrated optimal results when $k$ was set to 8. 
This variance reveals that the required intensity of strong augmentation techniques varies across different datasets, which may be attributed to differences in data distribution.

\section{Conclusion}
In this paper, we propose a semi-supervised segmentation framework for RS to address the challenge of limited labeled data. 
Our experimental results on mainstream RS datasets demonstrate that the proposed framework significantly improves segmentation accuracy compared to existing frameworks.

The primary contributions of this work include the development of a novel augmentation strategy, Uniform Strength Augmentation (USAug), which is specifically designed for unlabeled RS images to generate discrepancies, and an adaptive CutMix technique (AdaCM) that delicately enriches the information embedded in unlabeled RS images while mitigating confirmation bias.

While our method shows promising results, it is limited by the computational cost of extensive augmentations and the dependency on specific threshold values. 
Future research could explore the delicate design of the augmentation strength and adaptive thresholding mechanisms.

Overall, our findings provide a valuable contribution to the field of RS, offering a robust solution for scenarios with scarce labeled data and paving the way for further advancements in semi-supervised learning methodologies.

\bibliographystyle{splncs04}
\bibliography{ssl_bib.bib}

\end{document}